\pdfoutput=1

\documentclass[11pt]{article}

\PassOptionsToPackage{table}{xcolor}
\usepackage{ACL2023}

\usepackage{times}
\usepackage{latexsym}
\usepackage{booktabs}
\usepackage{linguex}
\usepackage{annotates}
\usepackage{array}

\usepackage[T1]{fontenc}

\usepackage{listings}
\usepackage{caption}
\usepackage{subcaption} 

\lstdefinelanguage{text}{
  moredelim=[s][\bfseries]{@@@}{:}
}

\usepackage[utf8]{inputenc}

\usepackage{microtype}
\usepackage{graphicx}
\usepackage{float}
\usepackage{multirow}

\usepackage{inconsolata}
\usepackage{adjustbox}

\title{Generalizability of Media Frames:  \\
Corpus creation and analysis across countries}

\author{
  Agnese Daffara\textsuperscript{1}, 
  Sourabh Dattawad\textsuperscript{1},
  Sebastian Padó\textsuperscript{1},
  Tanise Ceron\textsuperscript{2}\\
  \textsuperscript{1}Institute for Natural Language Processing, University of Stuttgart, Germany \\
  \textsuperscript{2}Bocconi University, Italy \\
  \texttt{\{agnese.daffara,pado\}@ims.uni-stuttgart.de} \\
  \texttt{tanise.ceron@unibocconi.it} \\
}

\begin{document}
\maketitle
\begin{abstract}

Frames capture aspects of an issue that are emphasized in a debate by interlocutors and can help us understand how political language conveys different perspectives and ultimately shapes people's opinions. The Media Frame Corpus (MFC) is the most commonly used framework with categories and detailed guidelines for operationalizing frames. It is, however, focused on a few salient U.S. news issues, making it unclear how well these frames can capture news issues in other cultural contexts. To explore this, we introduce \texttt{FrameNews-PT}, a dataset of Brazilian Portuguese news articles covering political and economic news  and annotate it
within the MFC framework. 
Through several annotation rounds, we evaluate the extent to which MFC frames generalize to the Brazilian debate issues. We further evaluate how fine-tuned and zero-shot models perform on out-of-domain data.
Results show that the 15 MFC frames remain broadly applicable with minor revisions of the guidelines. However, some MFC frames are rarely used, and novel news issues are analyzed using general 'fallback' frames. We conclude that cross-cultural frame use requires careful consideration.


\end{abstract}

\section{Introduction}

Frames are interpretation schemes used to organize reality by isolating and highlighting 
salient aspects of it \citep{otmakhova2024media}.
They combine
three levels: cognitive, (concerning the mental representations of the world),
semantic (concerning the linguistic structures involved); and communicative (their usage and impact on the audience). %

Frames are finding increasing application in  NLP  as a framework to analyze 
social media and news coverage. In particular the communicative dimension of frames  can be seen as a bridge between belief and action \citep{snow2005clarifying}. 
%
Following \citet{entman1993framing},  frames have four objectives: i) promote a particular definition of a problem; ii) promote an interpretation of it; iii) propose a moral evaluation; and iv) recommend a treatment or a solution. This makes them a valuable tool for identifying, measuring, and quantifying particular worldviews and communication strategies \citep{card2015media,mendelsohn-etal-2021-modeling}.

Frames are primarily a methodological framework, and not a fixed set of categories.
However, a widely used set is based on the Media Frame Corpus (MFC), 15 frames (see Table \ref{tab:frames_examples}) with annotation guidelines \citep{card2015media}, 
 developed to cover a limited number of issues concerning the U.S. context, e.g., gun control and climate change.

As in other NLP tasks, however, \textit{generalization} is an issue: coarse-grained frames risk failing to capture important aspects of a debate, notably ideological bias, while fine-grained frames may not generalize across debates. For example, \citet{mendelsohn-etal-2021-modeling} demonstrate that in news about migration, the generic frame \textit{cultural identity} shows only a slight tendency towards liberal texts.
Introducing  two related sub-frames \textit{hero: cultural diversity} and \textit{threat: national cohesion} leads to a clear separation by political ideology -- liberal in the former and conservative the latter case -- but these subframes are presumably irrelevant for other debates.
Similar questions arise in cross-lingual contexts: The MFC frames were also applied by SemEval 2023 Task 3 to 9 languages \cite{piskorski2023semeval}, but in an arguably  pragmatic setting where the applicability of the frames was not assessed, nor was inter-annotator agreement reported. 

Indeed, full cross-linguistic generalization of frames 
is far from clear, given previous work on the cross-lingual generalization of  semantic inventories \citep{EWN,reddy-etal-2017-universal,GILARDI18.11}. For full generalization, frames would show high \textit{agreement} -- their definitions and annotation guidelines would allow each markable in a new language to be assigned to a unique frame -- and \textit{completeness} -- for each instance in the new language, there would be a suitable frame. However, we do not have  empirical evidence so far.

Thus, we set out to answer these questions:
\begin{enumerate}\setlength{\itemsep}{0pt}\setlength{\topsep}{0pt}
    \item How high are agreement and completeness for MFC frames when annotating a representative sample of news articles despite having been originally developed on a small number of issues?
        \item How high are agreement and completeness of the MFC frames when applied to news reporting from other countries? 
    
            \item Even if the frames generalize well, is it possible to transfer computational models of frame prediction between languages?
    \end{enumerate}
 Our study proceeds by creating \texttt{FrameNews-PT}, a corpus of 300 news 
articles of 'hard news' \citep{doi:10.1177/1464884911427803} in Brazilian Portuguese.\footnote{The corpus will  be made available for the community upon acceptance. 
}
\texttt{FrameNews-PT} is based on the News Portal Recommendation dataset \citep{lucas2023npr}. It contains Brazilian news about, e.g., new regulations, taxes, and education.
We manually annotate the articles with MFC frames using a \textit{perspectivist}
approach \citep{cabitza2023toward}. 
%
We carry out several rounds of annotation to a) adapt the MFC guidelines while maintaining their level of detail, b) value the individual perspectives of annotators, and c) examine generalizability and overlap of frames (RQ1+2).

After corpus creation, we conduct modeling experiments (RQ3). We use two transformer-based models fine-tuned on the MFC and 9 chat-instructed models in a zero-shot setting, and analyze how well models perform out of domain on \texttt{FrameNews-PT} in comparison with the MFC.

Our results are generally promising: (Most) MFC frames are annotated with high agreement in Brazilian hard news, with little confusion among them. However, some frames appear rarely, being bound to U.S.-specific issues. Also, completeness is not perfect: Brazilian issues outside the scope of the original MFC  are often annotated with relatively generic frames as a fallback (e.g., \textit{Economic}), leading to some loss of information. On the modeling side, we find that zero-shot models perform better at frame prediction on the Brazilian data than transferred English classifiers, indicating that the data distributions are sufficiently dissimilar. Overall, use of MFC frames in new languages and issues calls for a reflected process.


\section{Related Work}
\label{sec:related-work}

Previous work in NLP has investigated how framing political language influences society and public opinion \citep{lecheler2015effects}, and how it shapes the spread of information \citep{gilardi2021policy} across  communication contexts, such as agenda-setting campaigns \citep{tsur2015frame, field2018framing} and news media \citep{card2015media}. Frames can help analyze argumentation, and frame detection can be seen as argument assessment \citep{lauscher2022scientia}. Given their connection to political ideologies \citep{alashri2015animates}, frames are also useful for analyzing and diversifying perspectives, e.g. in news recommendation \citep{mulder2021operationalizing}.

The task of frame detection has often been approached through topic modeling with unsupervised methods \citep{nguyen2013lexical, roberts2014structural, tsur2015frame, ajjour2019modeling}, which, however, tend to be highly corpus-specific. Other studies employ supervised classification techniques, including logistic regression \citep{card-etal-2016-analyzing}, neural networks \citep{naderi2017classifying, liu2019detecting}, and fine-tuning of pre-trained models \citep{kwak2020systematic}, all of which require robust gold standard annotations.

The Media Framing Corpus (MFC) \citep{card-etal-2016-analyzing}, building on the foundational work of \citet{boydstun2014tracking}, introduced a novel annotation framework that has become a reference for frame identification (cf. Section \ref{sec:data}). Several studies either use parts of the MFC or applied its guidelines to related datasets, e.g., \citet{kwak2020systematic} conduct a systematic analysis of the 15 frames in New York Times articles across 17 years.

\citet{khanehzar2019modeling} are the first to test the generalizability of MFC frames to other contexts. They apply the framework to Australian parliamentary speeches on same-sex marriage and migration, and fine-tuned three BERT-based classifiers. Their findings show a drop in accuracy when the models are applied across different contexts. It is unclear, however, if this is due to the failure of the frames to \textit{generalize}, or only to a shift in frame \textit{distribution}.

Other corpora and labeling schemes have been proposed, though they often focus on specific debate issues. Examples include the  Ballistic Missile Defense corpus \citep{morstatter2018identifying}, the Gun Violence Corpus \citep{liu2019detecting}, and the migration corpus by \citet{mendelsohn-etal-2021-modeling}.\footnote{For a comprehensive overview of framing detection and analysis, see \citet{ali2022survey} and \citet{otmakhova2024media}, offering a multidisciplinary perspective on the field.}

SemEval-2023 Task 3 introduced a sub-task on framing detection in online news within a multilingual setting \citep{piskorski2023semeval}. The dataset spans multiple topics and languages. It is annotated with an adaptation of the MFC framework \citep{piskorski2023news} and the proposed models, based on Transformer architectures, achieved strong performance. The focus of the study was on modeling, however, not on assessing frame generalizability.

\section{Data}
\label{sec:data}

\paragraph{The Policy Frames Codebook} An  key contribution to the  study of frames is the Policy Frames Codebook by \citet{boydstun2014tracking}, which introduced 15 framing dimensions applied to three debate issues: immigration, tobacco, and same-sex marriage. The guidelines for frame annotation in this codebook have been dynamically updated over time \citep{boydstun2014policy}. This work led to the creation of the Media Frames Corpus \cite{card2015media}, later expanded to include three additional issues: gun control, death penalty, and climate (MFC v4.0). The MFC was explicitly developed for three debate issues and the U.S. context, which is a potential limitation. Additionally, as noted by \citet{ali2022survey}, it may not distinguish sufficiently between issues and frames by introducing overly broad categories, such as \textit{Economic}.

For our study, we adopt the 15 framing dimensions as labeled in \citet{card2015media}, cf. full list in Appendix \ref{app:frames}. 
Our guidelines are a version of the Policy Frames Codebook, shortened from 45 to 9 pages to preserve the original information while making it more accessible. We remove specific edge cases and marginal notes, but our annotators are encouraged to consult the original guidelines in cases of doubt. Our version of the guidelines, in English, includes a description for each frame and the cues to recognize it. In addition, they include: i) a contextualization of the project, presenting the tool used for annotation (Google Sheets with drop-down menus), and ii) for each frame, an example drawn from MFC headlines on migration where annotators agreed on the frame (see Table \ref{tab:frames_examples}). 
In contrast to the
version used in SemEval 2023 \citep{piskorski2023news}, our guidelines retain all relevant content from the original codebook. The  guidelines will be available with the corpus.

\paragraph{Data collection} Our dataset consists of 300 
news articles randomly extracted from the News Portal Recommendation Corpus (NPR), a collection of 148,099 news articles in Portuguese from a Brazilian news portal called G1 with user click history \citep{lucas2023npr}. This dataset was not only chosen due to its potential application in news recommendations, but also because it covers a broad range of issues from a cultural and linguistic context different from the U.S. and the Global North generally. This allows us to investigate our research questions on the generalizability of framing annotations and model performance across news issues.

We select articles belonging to  the economics and politics sections of the newspaper. The articles in the NPR dataset often belong to more than one category, so they also span across other news categories as well. In this way, we ensure that articles i) consist of hard news, typically including topics such as foreign and domestic politics, economy and finance \citep{doi:10.1177/1464884911427803}, and ii) cover a broad domain, spanning multiple debate issues. A topic modeling analysis with BerTopic \cite{grootendorst2022bertopic} on NPR and \texttt{FrameNews-PT} shows that our selection of topics is representative of the whole NPR corpus (see Appendix \ref{app:news-categories}). To comply with model input length limitations, we limit our selection to articles shorter than 300 words and remove boilerplate (banners, ads).
\paragraph{Data annotation} We hired two annotators, both Master's students: one Brazilian and one Portuguese. We expect their cultural backgrounds to influence the interpretation and application of frames during the task. In order to acknowledge individual perspectives and subjectivity, we keep the annotations disaggregated and we evaluate the models on each annotator, adopting a perspectivist approach \citep{cabitza2023toward}. 

Like in the MFC, our annotators label the multiple body frames present in the body of the article. In contrast to the MFC, our annotations are only article-level, as we are not interested in span-level results. Then, they also choose the most prominent frame throughout the entire article, the \textit{primary frame}. In the MFC, the annotators also decide whether an article is relevant and do not carry out annotations on irrelevant articles.\footnote{A relevant article: i) concerns the policy issue in question, ii) takes place in or explicitly regards the United States, iii) has more than four lines, and iv) is a proper article, not a correction of a previous error or an obituary.} We skip this annotation phase for two main reasons: i) models are presented with both relevant and potentially irrelevant articles for classification, and we want the annotation process to align with the modeling setup; and ii) our dataset includes only hard news and has already been cleaned. Therefore, we aggregate irrelevant articles and those that do not match any frame's guidelines under a single label, namely \textit{Other}.
This frequently includes news on a personal rather than societal level, such as advice on personal finance, new commercial products, or celebrity news. An example of this category is the following headline, which describes a private decision by President Lula that does not match any frame dimension: 
\ex. 
    Lula decide não se mudar para a Granja do Torto antes da posse.\\
    \textit{Lula decides not to move to Granja do Torto before the inauguration.}

The annotation process was organized into four weekly rounds, with an increasing number of articles, based on the assumption that annotators become more efficient and familiar with the task over time: the first two rounds comprised 50 articles each, and the last two rounds contained 100 articles each. After each round, the annotators participated in a discussion session to address ambiguous cases and refined the application of the guidelines. Inter-annotator agreement (IAA) is measured using Krippendorff’s $\alpha$\footnote{Krippendorff's $\alpha$ was adopted following the MFC. It calculates agreement considering the whole pool of labels rather than individual frequency of use \citep{card2015media}.} after each round. We hypothesize that i) the IAA increases at each round because of the discussions, and ii) that the score before and after discussion becomes closer at each round.

\paragraph{Annotators' perspectives} In the discussion rounds, we observe the influence of the annotators' cultural backgrounds on the usage of frames. The Portuguese annotator struggled to understand certain contexts due to her different cultural perspective and she would sometimes need to look up acronyms, such as STF \textit{(Supremo Tribunal Federal)}. The Brazilian annotator, familiar with the local context, often had a deeper understanding of the reasons and potential consequences of news articles. Consider the following heading sentence, referring to scholarships for students:

\ex. 
    Reajuste de bolsas da Capes e do CNPq deve ser anunciado ainda em janeiro, diz Camilo Santana.\\
    \textit{Adjustment of Capes and CNPq scholarships should be announced in January, says Camilo Santana.}
    
The Brazilian annotator, aware that access to scholarships is a major issue in Brazil, considered the introduction of such a policy to be significant and labeled the article as \textit{Policy Prescription and Evaluation}. By contrast, the Portuguese annotator, having associated the term \textit{bolsas} with the stock exchange (\textit{bolsa de valores}), labeled it as \textit{Economic}.
These differences show that a perspectivist approach is helpful in annotation and modeling. We keep the annotations disaggregated in the evaluation.

\section{Analysis of Frame Generalization}
\label{sec:frame-annotations}

We can now investigate the generalization of the MFC frames to the Brazilian context.

\begin{figure}[tbh!]
    \centering    \includegraphics[width=0.9\linewidth]{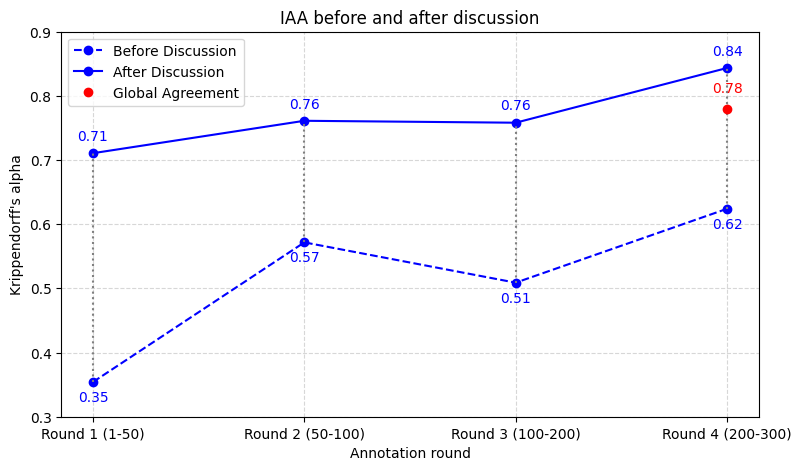}
    \caption{Inter-annotator agreement on the primary frame over four rounds, before and after discussion.}
    \label{fig:IAA}
\end{figure}

\paragraph{Inter-annotator agreement} As Figure \ref{fig:IAA} shows, 
the agreement score on the primary frame kept improving generally over the four rounds, with the exception of the third round, which contained more complex cases. The discussion was useful in distinguishing difficult frames. After the final revision, the global agreement on the dataset was $\alpha$=0.78. It is significantly higher than the one reported on the MFC by \cite{card2015media}, which never exceeds 0.70. 
This can be attributed to four main reasons:
i) in the MFC, new annotators with no previous knowledge were introduced in each round, potentially lowering the overall score; ii) discussing the guidelines with the same person, as we did, could in the worst case lead to overfitting (we alleviate the effects in modeling by keeping annotations disaggregated);
iii) annotators usually fell back to generic frames rather than specific ones; iv) they  used \textit{Other} to annotate articles about personal stories that did not match other frames in the guidelines, consequently raising the agreement.

\paragraph{IAA on single frames}

\begin{figure}[tb!]
    \centering   \includegraphics[width=1\linewidth]{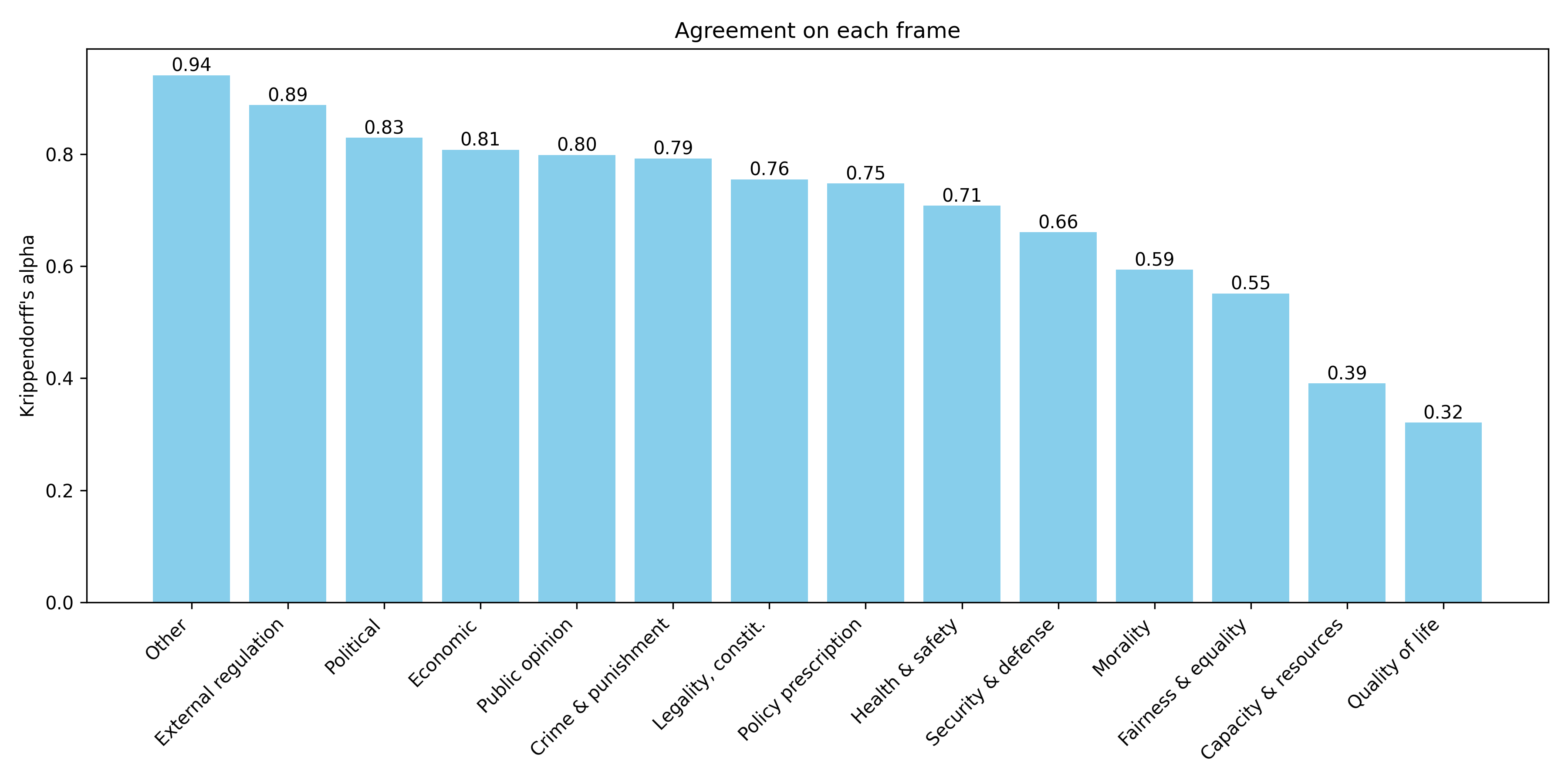}
    \caption{Agreement on each frame (primary).}    \label{fig:agreement_labels}
\end{figure}

We also compute agreement on single frames (Figure \ref{fig:agreement_labels}). The results confirm highest agreement for \textit{Other}, arguably due to its particular function in our annotations (see Section \ref{sec:data}).
Other frames with high scores include \textit{External Regulation and Reputation}, suggesting that this frame is particularly well defined, \textit{Political}, and \textit{Economic}, which is not surprising given that these are the dominant categories in the dataset. On the other hand, the frames with the lowest agreement are \textit{Capacity and Resources} and \textit{Quality of Life}. 

\begin{figure}[h!]    \includegraphics[width=1\linewidth]{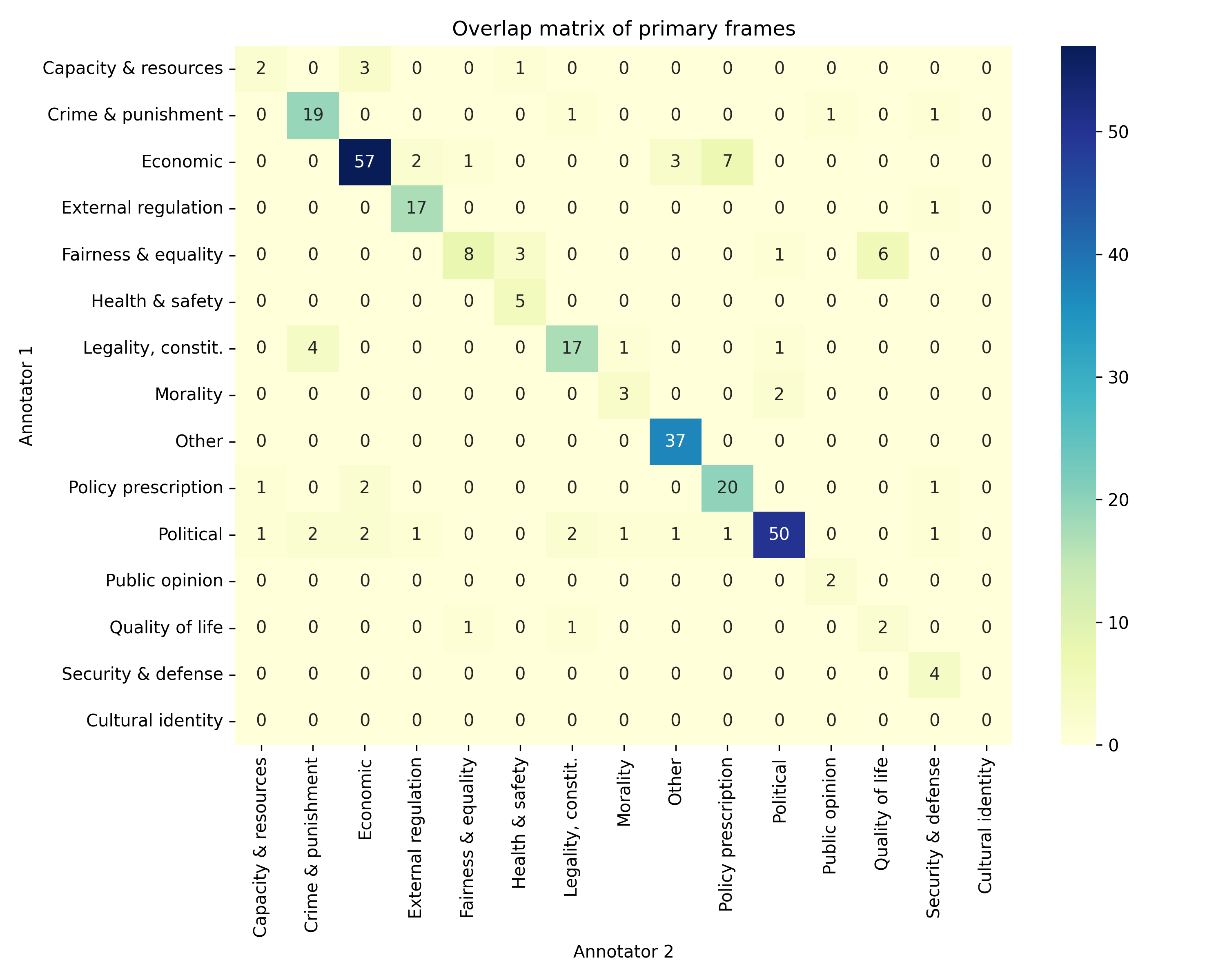}
    \centering
    \caption{Primary frames overlap between annotators.}
    \label{fig:overlap}
\end{figure}

\paragraph{Frame overlap}
Figure \ref{fig:overlap} shows the overlap between the two annotators on primary frames, indicating which frame pairs struggle with high agreement. These include: i) \textit{Crime and Punishment} vs. \textit{Legality, Constitutionality and Jurisprudence} ii) \textit{Fairness and Equality} vs. \textit{Quality of Life}, iii) \textit{Policy Prescription and Evaluation} vs. \textit{Economic}.

\textit{Quality of Life} and \textit{Fairness and Equality} frequently co-occur when discussing social issues, as in the following:

\ex.
    Ministério diz que novo Bolsa Família terá R\$ 18 bilhões para crianças de até 6 anos. \\
    \textit{The Ministry says that the new Bolsa Família will have R\$ 18 billion for children up to 6 years old}.

\textit{Crime and Punishment} and \textit{Legality, Constitutionality and Jurisprudence} overlap when talking about legal measures to punish someone,
while \textit{Policy Prescription} and \textit{Economic} occur together when speaking of policies that affect economy. 

We would argue that this overlap is not due to failure of a frame to generalize. Rather, it 
is a case of a complex situation
which is described as having an import on multiple aspects. Such multi-class cases are intrinsic to framing.

\begin{figure*}[h!]
    \centering
    \includegraphics[width=1\linewidth]{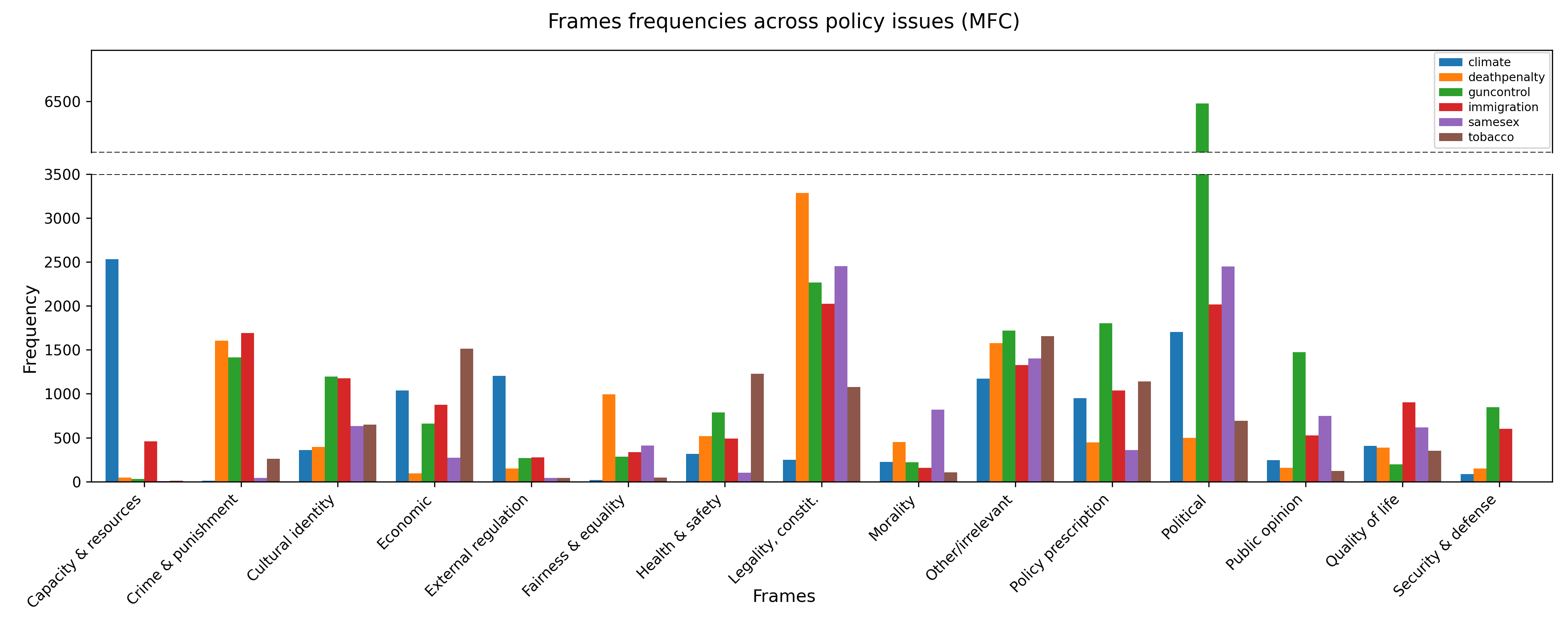}
    \caption{Frames frequencies across policy issues in the MFC.}
    \label{fig:freq-mfc}
\end{figure*}

\begin{figure}[tbh!]
    \centering
    \includegraphics[width=1\linewidth]{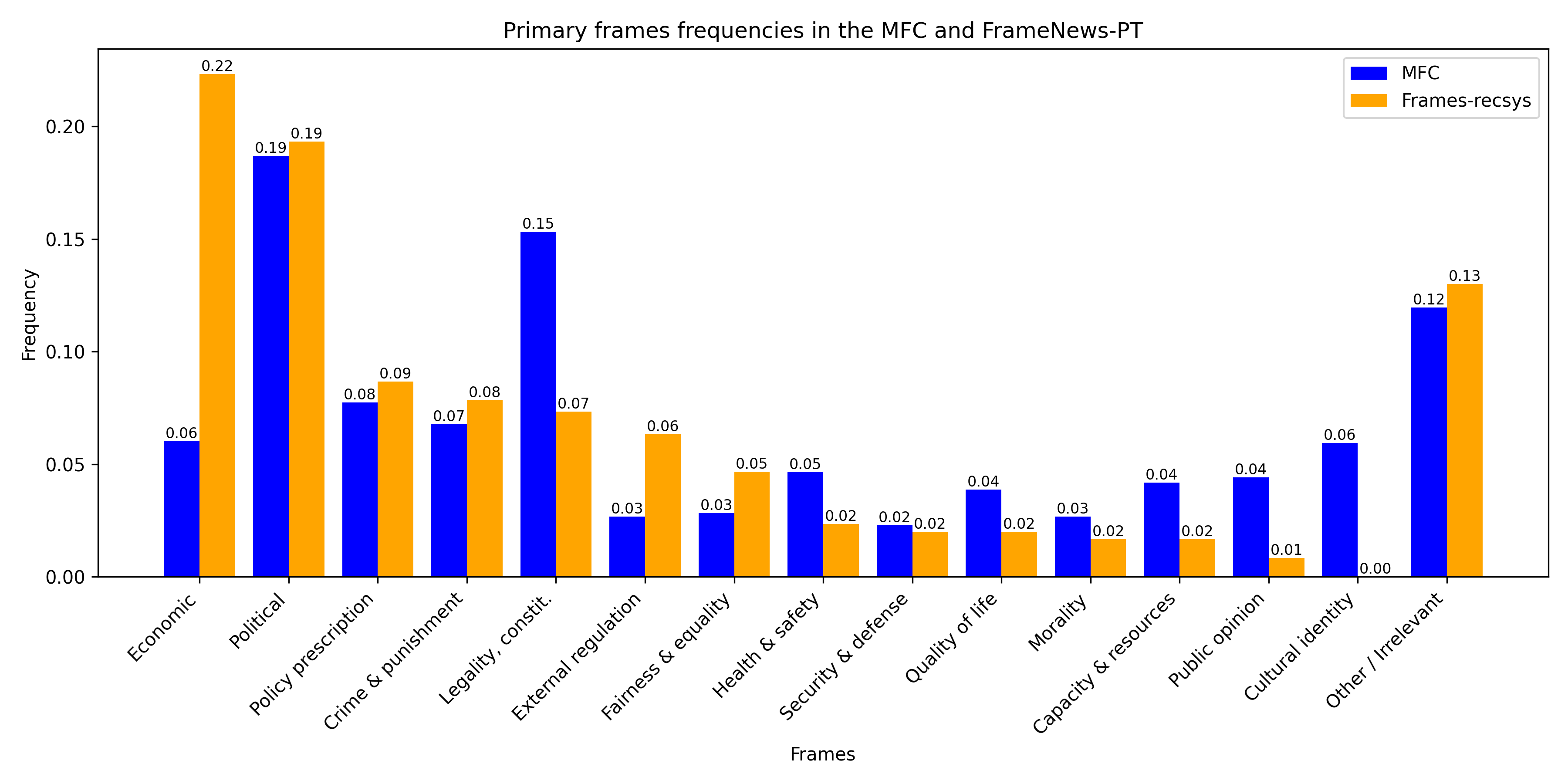}
    \caption{Primary frames frequency in the MFC compared to \texttt{FrameNews-PT} (normalized).}
    \label{fig:labels_freq}
\end{figure}

\paragraph{Completeness: Frame frequency} 
While a low frequency does not  imply that a frame does not generalize, high frequencies for \texttt{Other} or general frames may indicate lack of completeness. We therefore compare  frame frequencies between  MFC and \texttt{FrameNews-PT} (Figures \ref{fig:freq-mfc} and \ref{fig:labels_freq}).

Some frames, like \textit{Economic}, \textit{Fairness and Equality}, and \textit{External Regulation and Reputation} appeared more frequently in \texttt{FrameNews-PT}, suggesting they transfer well to Brazilian debate coverage. In contrast, six frames were used fewer than 20 times: \textit{Health and Safety}, \textit{Security and Defense}, \textit{Quality of Life}, \textit{Capacity and Resources}, \textit{Morality}, and \textit{Public Opinion}. We hypothesize that these frames are not used frequently because they are tied to specific issues in the MFC; for example, \textit{Capacity and Resources} is strongly associated to climate change. \textit{Cultural Identity} was never used by our annotators, which appears due to its specific use in the U.S. issues migration and gun control. 

The proportion of \textit{Other} in our corpus is almost the same as in the MFC (13\%), although it is used more broadly in our annotations (see Section \ref{sec:data}). This means that, even though we sample from a much broader range of issues, the 15 frames still suffice to classify the news. However, annotators in our case tended to fall back to more general frames, indicating a lack of completeness.

\paragraph{Annotator survey}
To validate our interpretations from above, we asked the annotators to fill a survey about the annotation process, and discussed the results in a meeting. The feedback from the survey confirmed that annotators found our summarized guidelines easier to read and understand than the original one by \citet{boydstun2014tracking}. They identified two main limitations. First, some frame descriptions are overly specific to the U.S. context. For example, \textit{Security and Defense} included immigration issues at the U.S. border (cf. Appendix \ref{app:frame-descriptions}), a topic not prominent in Brazil. Second, some frames are not described  sufficiently distinctly by the guidelines, leading to overlap. 

Regarding the frequency of frames, the annotators attributed the low usage of certain frames to the topics covered, confirming our hypothesis that some frames are specific to particular issues. However, when asked whether they would add or eliminate any category, they agreed that the same tagset could be used for future tasks, provided that some adjustments are made, making the frames more context-aware and contrastive. 

The articles tagged as \textit{Other} mainly focused on private issues rather than on societal-level concerns, which is why they did not match any frame. While these could be covered by  new frames (e.g., \textit{Personal Finance}, \textit{Advertisement}), adding such fine-grained categories would be misaligned with the goals of framing analysis for hard news.

\paragraph{Conclusions}
In sum, we find that the 15 frames generalize well enough that they can be employed in future comparable projects, with the following revisions to the guidelines:
i) replace U.S.-specific examples with locally relevant counterparts (e.g., substitute \textit{Medicare} with \textit{Bolsa Família});
ii) clarify contrasts between overlapping frames based on the local context (e.g., \textit{Quality of Life} vs. \textit{Health and Safety});
iii) better define the scope of \textit{Other}.

\section{Frame Prediction on \texttt{FrameNews-PT}}
\label{sec:methods}

\begin{table*}[h!]
\small
\centering
\begin{tabular}{lll>{\columncolor{gray!20}}l>{\columncolor{gray!20}}lll>{\columncolor{gray!20}}l}
\toprule
& \textbf{Model} & \textbf{Dataset} & \multicolumn{1}{l}{\textbf{Accuracy}} & \multicolumn{1}{l}{\textbf{Global F$_1$}} & \multicolumn{1}{l}{\textbf{F$_1$ (ann 1)}} & \multicolumn{1}{l}{\textbf{F$_1$ (ann 2)}} & \multicolumn{1}{l}{\textbf{Cohen's $\kappa$}} \\
\midrule
\multicolumn{2}{l}{\multirow{2}{*}{Majority class baseline}} & MFC & 0.21 & - & - & - & - \\
& & \texttt{FrameNews-PT} & 0.23 & - & - & - & - \\
\midrule
\multirow{2}{*}{Fine-tuned} & \multirow{2}{*}{XLM-RoBERTa} & MFC & \textbf{0.67 $_{\pm 0.00}$} & \textbf{0.68 $_{\pm 0.00}$} & - & - & - \\
& & \texttt{FrameNews-PT} & 0.53 $_{\pm 0.00}$ & 0.48 $_{\pm 0.01}$ & 0.49 $_{\pm 0.00}$ & 0.47 $_{\pm 0.01}$ & - \\
\midrule
\multirow{2}{*}{Fine-tuned} & \multirow{2}{*}{Multilingual-E5} & MFC & 0.66 $_{\pm 0.00}$ & 0.67 $_{\pm 0.01}$ & - & - & - \\
& & \texttt{FrameNews-PT} & 0.56 $_{\pm 0.01}$ & 0.50 $_{\pm 0.01}$ & 0.51 $_{\pm 0.01}$ & 0.49 $_{\pm 0.01}$ & - \\
\midrule
\multirow{2}{*}{Zero-shot} & \multirow{2}{*}{Qwen2.5-7B-Instruct} & MFC & 0.40 $_{\pm 0.10}$ & 0.38 $_{\pm 0.01}$ & - & - & 0.50 $_{\pm 0.00}$ \\
& & \texttt{FrameNews-PT} & 0.53 $_{\pm 0.02}$ & 0.44 $_{\pm 0.02}$ & 0.45 $_{\pm 0.03}$ & 0.42 $_{\pm 0.03}$ & 0.49 $_{\pm 0.18}$ \\
\midrule
\multirow{2}{*}{Zero-shot} & \multirow{2}{*}{gpt-4o-2024-08-06} & MFC & 0.46 $_{\pm 0.00}$ & 0.46 $_{\pm 0.00}$ & - & - & 0.69 $_{\pm 0.00}$ \\
& & \texttt{FrameNews-PT} & \textbf{0.59 $_{\pm 0.01}$} & \textbf{0.50} $_{\pm 0.03}$ & 0.51 $_{\pm 0.04}$ & 0.49 $_{\pm 0.03}$ & 0.75 $_{\pm 0.06}$ \\
\bottomrule
\end{tabular}
\caption{Results of the models on the two datasets. For XLM-RoBERTa and Multi-E5, standard deviations are across random seeds. For Qwen and ChatGPT, they are across templates. Accuracy considers the output as correct if it matches at least one annotation. Global F$_1$ is calculated on the Gold Standard for the MFC, while it is averaged on both annotators for \texttt{FramesNews-PT}. Cohen's $\kappa$ measures reliability across prompt templates. Best results for encoder-based and generative models are \textbf{in bold}.}
\label{tab:classification-results}
\end{table*}

Even very good frame generalizability does not imply that frame prediction models transfer well across languages, and at least for some frames we see substantial shifts in frequency (cf. Figure \ref{fig:labels_freq}). Therefore, a) we run two classifiers trained on the MFC corpus, testing them  both on MFC and on \texttt{FrameNews-PT}; b), we classify \texttt{FrameNews-PT} in a zero-shot setting with 8 chat-instructed LLMs ranging from 1B to 12B parameters as well as ChatGPT-4o (exact parameter size unknown). All chosen models are multilingual since \texttt{FrameNews-PT} is in Portuguese.  

\subsection{Experimental Setup}
\label{sec:experimental-setup}


We  split the Media Frames Corpus (MFC) into 70\% training (21,751 examples), 10\% validation (3,107), and 20\% test (6,214). The test set is held out and only used for final evaluation. All splits are fixed across runs to ensure consistency. The \texttt{FrameNews-PT} corpus is always used as a test set, in order to understand how classifiers perform on a novel set of news issues from a different country.

\paragraph{Supervised classifiers} We fine-tune two pretrained multilingual language models: XLM-RoBERTa-base \cite{DBLP:journals/corr/abs-1911-02116} and Multilingual-E5-base \cite{wang2024multilingual} for single-label classification over 15 frames. Inputs are tokenized using the model tokenizers with a maximum sequence length of 512, and models are trained using cross-entropy loss.
Each model is trained for up to 10 epochs with early stopping based on validation loss, using a patience of 3. We use a batch size of 64 for both training and evaluation. The learning rate is 5e-5. All experiments are repeated with five different random seeds. 

\paragraph{Zero-shot classifiers} For zero-shot classification, we prompt multilingual chat-instructed models from the Qwen \citep{qwen2024qwen2}, Google Gemma \citep{team2025gemma}, Llama \citep{grattafiori2024llama}, and GPT \citep{achiam2023gpt} families. Our prompts include the article to be labeled, the shortened annotation guidelines, and a task description. We evaluate the models on 3 prompt templates (see examples in \ref{appendix:prompt-templates}) and temperature 0.

\paragraph{Evaluation metrics}
We evaluate the performance of the models on accuracy and F1-score. Accuracy is measured taking the \textit{perspectivist} approach into account.
That is, we consider the answer of the model correct when it matches the primary labels of one of the annotators. The F1-score for the MFC corpus is based on the gold labels while we report the F1-scores for each annotator in  \texttt{FrameNews-PT} separately. We also calculate Cohen's $\kappa$ for checking the reliability of the models answers across the evaluated prompt templates.

\subsection{Results}
\label{sec:discussion}

\begin{figure*}[h!]
    \centering
    \includegraphics[width=0.8\linewidth]{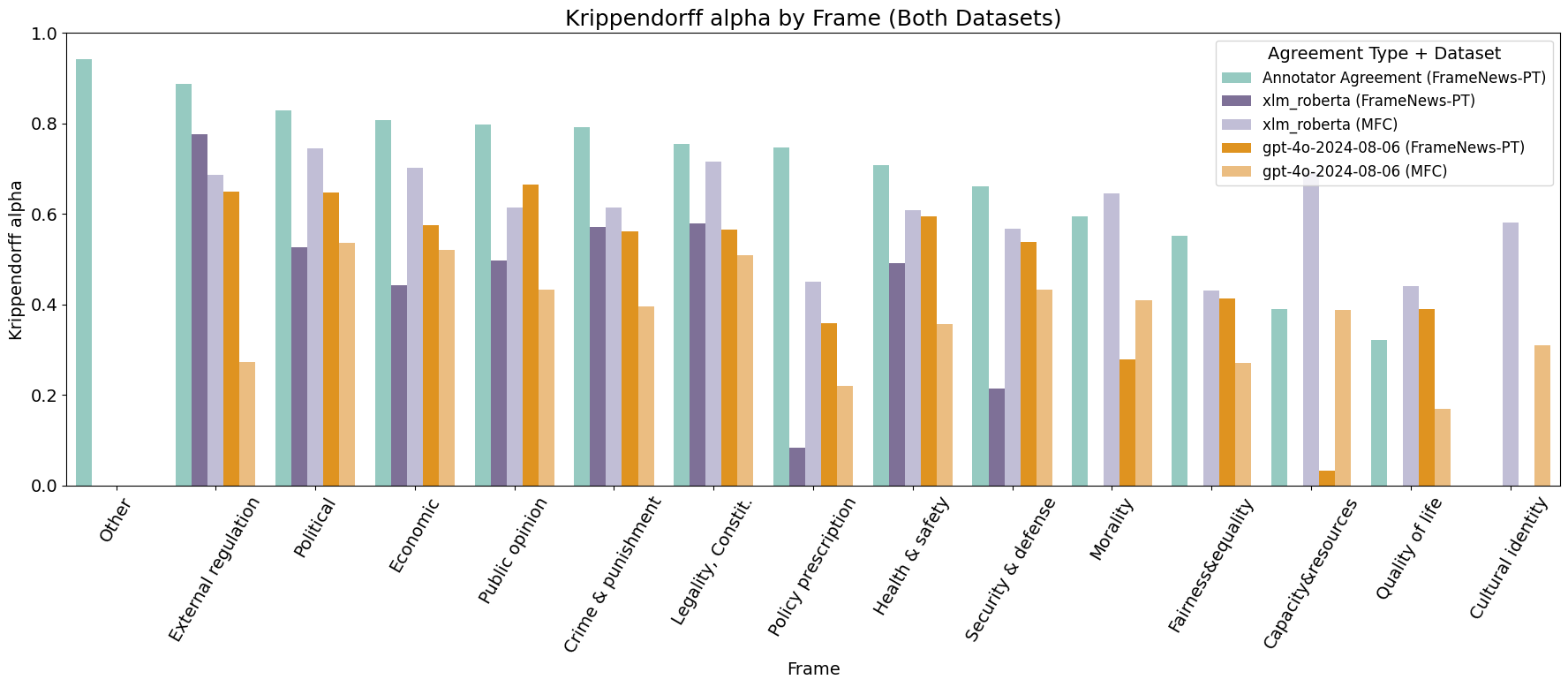}
    \caption{Agreement between models' predictions and Annotator 1 on \texttt{FrameNews-PT} and  MFC, and Agreement between annotators on \texttt{FrameNews-PT}.}
    \label{fig:agreement-models}
\end{figure*}

We report results for the two supervised models and the two best-performing zero-shot models in Table~\ref{tab:classification-results}; for the
remaining results see the Appendix. 

The best model among the supervised learning classifiers for MFC is \texttt{RoBERTa-XLM} (acc=67\%) while for \texttt{FrameNews-PT} it is Multi-E5 (acc=56\%). \texttt{RoBERTa-XLM} clearly outperforms the zero-shot classifiers on the MFC. On the other hand, ChatGPT-4o is the best model among the zero-shot classifiers for MFC (acc=46\%) and \texttt{FrameNews-PT} (acc=59\%) -- not surprisingly, given its parameter count. Note that when comparing ChatGPT-4o and RoBERTa-XLM, ChatGPT-4o yields the best result on \texttt{FrameNews-PT}, while RoBERTa-XLM outperforms it by 8 points accuracy on the MFC. The F1-score results follow similar patterns. 

These findings suggest a trade-off: supervised models trained on annotated data achieve strong in-domain performance but are less effective in transfer to new corpora, whereas large zero-shot models offer more robust cross-domain performance but fall short in accuracy when compared to supervised models in an in-domain setup.

As for reliability across prompt templates, larger models are more reliable \citep{ceron-et-al-2024}: ChatGPT-4o achieves 0.75 and 0.69 Cohen's $\kappa$ on the MFC and \texttt{FrameNews-PT} respectively, whereas Qwen2.5-7B reaches only 0.46 and 0.49.

Figure \ref{fig:agreement-models} shows model performance and inter-annotator agreement across frames. 
The agreement between human annotators is always higher than model-human agreement, except for \textit{Quality of Life}. The models struggle the most with the category \textit{Other} -- for the supervised models, this is presumably because this category is very infrequent in the MFC; for the zero-shot models, we hypothesize that this is due to its status as a 'catch-all' category which is hard to learn.

While supervised models perform well to relatively well on all frames on the MFC, they struggle to generalize to \texttt{FrameNews-PT}, particularly for \textit{Policy Prescription and Evaluation}, \textit{Morality}, \textit{Fairness and Equality}, \textit{Capacity and Resources}, and \textit{Quality of Life}. The latter three  also  have the lowest inter-annotator agreement.
Our best zero-shot model, GPT-4o, performs worst for \textit{Policy Prescription and Evaluation} and \textit{Quality of Life} on the MFC, arguably due to blurred boundaries of these frames
 (cf. Section \ref{sec:frame-annotations}). On \texttt{FrameNews-PT}, it struggles the most with \textit{Morality} and \textit{Capacity and Resources}, likely due to the shift in topics.

These observations give rise to a final question: is the 'limiting factor' in the models' performance on \texttt{FrameNews-PT} data quality on the MFC, on \texttt{FrameNews-PT}, or the shift in distribution? To understand this, we ran a linear regression to predict  GPT-4o's accuracy on the MFC at the frame level (cf. Figure \ref{fig:agreement-models}).
Using three predictors: IAA on MFC, IAA on \texttt{FrameNews-PT}, the regression explains 53\% of the variance, with only the IAA on the MFC being a significant predictor at p$<$0.05. This is surprising, since GPT-4o is not even trained on the MFC. Our take on this result is that the MFC inventory includes frames with vague or otherwise deficient definitions, which leads to problems both in terms of annotation (low IAA) and in terms of prediction (low performance).


\section{Conclusion}
\label{sec:conclusion}


This study introduces \texttt{FrameNews-PT}, the first dataset of Brazilian Portuguese news with both users' click history (sampled from the NPR corpus) and annotation with the 15 Media Frames Corpus frames. It  offers a valuable resource for both the NLP and recommender systems communities. 

Our analysis in this paper has focused on the generalizability of the MFC frames to the Brazilian context. We find generally high agreement among annotators, which we interpret as a promising sign; however, the MFC frames are not complete. In the absence of specific frames annotators tend to fall back to general frames, with reduced specificity as consequence. This trend is largely due to the original frames and annotation guidelines being developed for use within the specific U.S. context.

Our modeling experiments show that zero-shot generative models achieve robust results on our dataset.
However, the very good in-domain results for the supervised MFC models indicate that annotated data is still pivotal to reach very good performance; both types of models suffer from shortcomings in  frame definition and data creation.

We conclude that the 
15 MFC frames remain broadly applicable and continue to capture essential dimensions of political discourse, even across countries and languages. To improve generalization, we recommend refining the annotation guidelines to offer clearer distinctions between overlapping categories and incorporating locally relevant examples (e.g., \textit{Bolsa Família} instead of \textit{Medicare}). Such refinements would enhance the framework’s adaptability to diverse cultural contexts and evolving policy debates.

\section*{Limitations}
The main limitation of our study is the dimension of our dataset. We had to limit the number of articles to 300, which is a small sample size. The main reason is the expense to run this type of study. Annotating frames in news articles is costly because the guidelines are extensive to learn and the news articles are long. Annotators also need time to choose all the body frames present in the articles, and then on the primary frame for the entire article. Because of the same reason, we have only chosen articles from a Brazilian news outlet, which may limit the generalizability of our results.

\section*{Ethics Statement}

The data used in this study is publicly available and was collected from the News Portal Recommendation Corpus (NPR) \citep{lucas2023npr}. The corpus contains news articles from Brazilian outlets, and the annotation process was conducted by two Master's students who were compensated for their work. The study adheres to ethical guidelines for research involving human participants, ensuring that the annotators were informed about the purpose of the study and their rights as participants. The research does not involve any sensitive or personal data, and all data used in the study is anonymized.


\bibliographystyle{acl_natbib}
\bibliography{custom,anthology}

\newpage
\clearpage

\appendix

\section{Appendix}
\label{sec:appendix}

\subsection{Frames}
\label{app:frames}

We adopt the labels from \citet{card2015media}. Note that some names from \citet{boydstun2014tracking} were changed. For example, \textit{Constitutionality and Jurisprudence} was modified to \textit{Legality, Constitutionality and Jurisprudence}, while \textit{Law and Order, Crime and Justice} was changed into \textit{Crime and Punishment}. In the updated Codebook \cite{boydstun2014policy}, the names of some labels were adapted again (e.g., \textit{Morality} is called \textit{Morality and Ethics}).

\begin{table}[ht!]\small
\centering
\footnotesize
\begin{tabular}{p{3cm} p{4cm}}
\toprule
\textbf{Frame} & \textbf{Example} \\ \hline
Capacity and Resources & \textit{Immigration debate: Illegals take jobs from Americans} \\ \hline
Crime and Punishment & \textit{Two charged in deaths of illegal immigrants in truck} \\ \hline
Cultural Identity & \textit{Ethnic shift: Immigration—an Irish enclave learns a new language; Mexican immigrants boost a growing Latino population} \\ \hline
Economic & \textit{Society makes no-interest loans to New York's immigrants} \\ \hline
Fairness and Equality & \textit{Strict immigration law unfairly targets Hispanics} \\ \hline
External Regulation and Reputation & \textit{‘International village’ gets hostile reception} \\ \hline
Health and Safety & \textit{Colombian drug violence leads to exodus} \\ \hline
Legality, Constitutionality and Jurisprudence & \textit{House approves bill to abolish INS; The Senate will begin work next week on its own measure dealing with the immigration agency} \\ \hline
Morality & \textit{County's undocumented workers say they aren't here to ‘steal’} \\ \hline
Other & \textit{U.S. under pressure to carry bigger load} \\ \hline
Policy Prescription and Evaluation & \textit{President Donald Trump stalls on promise to eliminate J-1 visa program} \\ \hline
Political & \textit{Following Trump voter fraud allegations, claim that 5.7 million non-citizens voted is wrong} \\ \hline
Public Opinion & \textit{Immigration: Political evangelicals feel push to take sides} \\ \hline
Quality of Life & \textit{Big money, cheap labor} \\ \hline
Security and Defense & \textit{Decision on refugees overdue; U.S. officials must loosen immigration restrictions} \\ \bottomrule
\end{tabular}
\caption{The 15 frames and their examples extracted from the MFC headlines on migration.}
\label{tab:frames_examples}
\end{table}

\subsection{\texttt{FrameNews-PT} and NPR analysis}

For the NPR dataset, the average number of words in article titles is $14.9 \pm 3.6$, while the average number of words in article bodies is $471.7 \pm 391.9$ \citep{lucas2023npr}. Table \ref{tab:comparison_topics} shows the most frequent topics in NPR and \texttt{FrameNews-PT}.

\label{app:news-categories}


\begin{table*}[h!]
\centering
\small
\setlength{\tabcolsep}{6pt} 
\renewcommand{\arraystretch}{1.3} 

\begin{tabular}{l p{7cm} p{7cm}}
\hline
\textbf{id} & \textbf{\texttt{FrameNews-PT}} & \textbf{NPR} \\
\hline
0 & \cellcolor{yellow!30} presidente, lula, bolsonaro, governo, ex, feira, nesta, petrobras, conselho, pt & \cellcolor{yellow!30} lula, presidente, governo, ex, petrobras, conselho, feira, ministro, cargo, nesta \\
1 & \cellcolor{orange!30} governo, haddad, sobre, fiscal, tributária, imposto, vai, ministro, câmara, texto & \cellcolor{yellow!30} bolsonaro, presidente, federal, atos, stf, tribunal, ex, ministro, república, justiça \\
2 & \cellcolor{cyan!30} federal, atos, stf, tribunal, prisão, justiça, torres, ministro, polícia, pf & \cellcolor{orange!30} governo, haddad, fiscal, sobre, tributária, proposta, câmara, regra, ministro, imposto \\
3 & \cellcolor{orange!30} mínimo, bilhões, inflação, salário, 2023, pib, governo, valor, anno, tesouro & \cellcolor{orange!30} juros, taxa, brasil, tesouro, dívidas, bancos, milhões, bilhões, inss, estados \\
4 & \cellcolor{orange!30} dívidas, bancos, tesouro, milhões, bilhões, empréstimos, governo, juros, estados, taxa & \cellcolor{green!30} café, rural, vídeos, bacalhau, globo, produtos, soja, onde, vem, produção \\
5 & \cellcolor{violet!30} empresa, petrobras, bahia, companhia, vendas, marca, varejo, ford, refinaria, unilever & \cellcolor{violet!30} empresa, petrobras, bahia, companhia, vendas, marca, ford, varejo, refinaria, bilhões \\
6 & \cellcolor{green!30} rural, bacalhau, globo, vídeos, soja, produtos, saiba, água, assistidos, café & \cellcolor{blue!30} mínimo, inflação, salário, bilhões, 2023, pib, governo, anno, reajuste, família \\
7 & \cellcolor{blue!30} brasil, auxílio, divórcios, mulheres, milhões, número, julho, ativos, datafolha, analisa & \cellcolor{red!30} microsoft, internet, telemarketing, explorer, uso, bloqueadores, partir, clienti, codice, applicazione \\
8 & \cellcolor{red!30} microsoft, telemarketing, internet, explorer, ataque, clientes, secretaria, código, aplicativo, partir & \cellcolor{blue!30} crédito, cartão, americanas, causas, trabalho, bancos, linha, trabalhadores, diagnóstico, câmbio \\
\hline
\end{tabular}

\caption{Comparison between top topics in \texttt{FramesNews-PT} and NPR, detected with BERTopic \cite{grootendorst2022bertopic}. Topics: 
\colorbox{yellow!30}{\rule{0.1cm}{0.1cm}} politics, 
\colorbox{orange!30}{\rule{0.1cm}{0.1cm}} economics, 
\colorbox{cyan!30}{\rule{0.1cm}{0.1cm}} justice, 
\colorbox{green!30}{\rule{0.1cm}{0.1cm}} production, 
\colorbox{violet!30}{\rule{0.1cm}{0.1cm}} companies, 
\colorbox{red!30}{\rule{0.1cm}{0.1cm}} technology,
\colorbox{blue!30}{\rule{0.1cm}{0.1cm}} social support.
}
\label{tab:comparison_topics}
\end{table*}

\subsection{Expenses}

We spent 96 dollars for evaluating ChatGPT-4o with the OpenAI API (3 prompt templates x 6,214 = 18,642 prompts). We paid 445 euros for each annotator to annotate the 300 examples in 27 hours of work. 

\subsection{Prompt templates}
\label{appendix:prompt-templates}

\begin{lstlisting}[caption={Prompt template without guidelines: zero1}]
### PROMPT:
"{content}"

### TASK:
Classify the PROMPT above into exactly ONE of these frame categories:

"1" Economic
"2" Capacity and resources
"3" Morality
"4" Fairness and equality
"5" Legality, Constitutionality, Jurisdiction
"6" Crime and punishment
"7" Security and defense
"8" Health and safety
"9" Quality of life
"10" Cultural identity
"11" Public opinion
"12" Political
"13" Policy prescription and evaluation
"14" External regulation and reputation
"15" Other

### CLASSIFICATION GUIDELINES:

{guidelines}

Base your answer only on the PROMPT and the guidelines provided above. Answer as a single number ("1", ..., "15") corresponding to the most appropriate category.

### ANSWER:
\end{lstlisting}

\begin{lstlisting}[caption={Prompt template with guidelines: The guidelines can be found in the github link. The prompt }]
### CLASSIFICATION GUIDELINES: 

{guidelines}

### PROMPT:
"{content}"

### TASK:
Classify the PROMPT above into exactly ONE of the categories below. 
"1" Economic
"2" Capacity and resources
"3" Morality
"4" Fairness and equality
"5" Legality, Constitutionality, Jurisdiction
"6" Crime and punishment
"7" Security and defense
"8" Health and safety
"9" Quality of life
"10" Cultural identity
"11" Public opinion
"12" Political
"13" Policy prescription and evaluation
"14" External regulation and reputation
"15" Other

 Answer as a single number ("1",..., "15") corresponding to the most appropriate category. 
 ### ANSWER:
 
\end{lstlisting}

\subsection{Description of frames}
\label{app:frame-descriptions}
\small

Here we provide the descriptions of frames in our guidelines.


\paragraph{1. Economic:} The costs, benefits, or any monetary/financial implications of the issue (to an individual, family, organization, community or to the economy as a whole). Can include the effect of policy issues on trade, markets, wages, employment or unemployment, viability of specific industries or businesses, implications of taxes or tax breaks, financial incentives, etc.

\paragraph{2. Capacity and resources:}The lack or availability of resources (time, physical, geographical, space, human, and financial resources). The capacity of existing systems and resources to carry out policy goals. The easiest way to think about it is in terms of there being "not enough" or “enough” of something. The capacity or resources may be an impediment to solving a problem or adequately addressing an issue.

\paragraph{3. Morality:} Any perspective that is compelled by religious doctrine or interpretation, duty, honor, righteousness or any other sense of ethics or social or personal responsibility. It is sometimes presented from a religious perspective (i.e. “eye for an eye”), but non-religious frames can also be used. For example, the moral imperatives to help others can be used to justify military intervention or foreign aid, social programs such as Medicare, welfare, and food stamps. Appeals that a policy move “is just the right thing to do” or “would indicate a recognition of our shared humanity” may reflect humanist morality. The commitment aspect of marriage would evoke feelings of morality. Environmental arguments that focus on responsible stewardship or “leaving something for our children” are based in a sense of responsibility or morality. Lawbreakers, including illegal immigrants, can be presented as fundamentally immoral, conversely breaking a law that is bad or unjust can be presented as moral (e.g., Rosa Parks). Enacting protective legislation, such as laws that protect children from pedophiles, guns, violence, poverty, or failure to do so can also be presented using moral frames.

\paragraph{4. Fairness and equality:} The fairness, equality or inequality with which laws, punishment, rewards, and resources are applied or distributed among individuals or groups. Also the balance between the rights or interests of one individual or group compared to another individual or group. Fairness and Equality frame signals often focus on whether society and its laws are equally distributed and enforced across regions, race, gender, economic class, etc. Many gender and race issues, in particular, include equal pay, access to resources such as education, healthcare and housing. Another example could be fairness considerations about whether punishments are proportional to crimes committed. The frame is also used when discussing social justice, discrimination and talk of an inmate’s innocence or exogeneration.

\paragraph{5. Legality, Constitutionality, Jurisdiction:} The legal, constitutional, or jurisdictional aspects of an issue. Legal aspects include existinglaws, reasoning on fundamental rights and court cases; constitutional aspects include all discussion of constitutional interpretation and/or potential revisions; jurisdiction includes any discussion of which government body should be in charge of a policy decision and/or the appropriate scope of a body’s policy reach. This frame deals specifically with the authority of government to regulate, and the authority of individuals/corporations to act independently of government. Of special note are constraints imposed on freedoms granted to individuals, government, and corporations via the Constitution, Bill of Rights and other amendments. Some frequent arguments and issues are: i) the right to bear arms; ii) equal protection; iii) free speech and expression; iv) the constitutionality of restricting individual freedoms and imposing taxes; v) conflicts between state, local or federal regulation and authority, or between different branches of government; vi) legal documentation (green card, visas, passports, driver licenses, marriage license, etc.).

\paragraph{6. Crime and punishment:} The violation of policies and its consequences. It includes enforcement and interpretation of civil and criminal laws, sentencing and punishment with retribution or sanctions. This frame includes: i) deportation when an individual does not have the necessary documents that grant legal standing; ii) increases or reductions in crime; iii) punishment and execution; iv) resources analysis like DNA analysis. Usually found together with other frames, such as Economic, Legality, constitutionality and jurisdiction, Morality, and Capacity and resources. The primary frame should be chosen according to where the emphasis is.

\paragraph{7. Security and defense:}
Any threat to a person, group, or nation, or any defense that needs to be taken to avoid that
threat.
Security and Defense frames differ from Health and Safety frames in that Security and Defense
frames address a preemptive action to stop a threat from occurring, whereas Health and Safety
frames address steps taken to ensure safety in the event that something happens. It can include
efforts to build a border fence or “secure the borders,” issues of national security including
resource security, efforts of individuals to secure homes, neighborhoods or schools, and efforts
such as guards and metal detectors that would defend children from a possible threat. Discussion
regarding terrorist activity should be coded as Security and Defense (e.g. arrests of terrorists,
immigrants linked to terrorism activity, increased border security to prevent terrorism). Arrests at
the border will receive both a Crime and Punishment and Security and Defense frame but the
primary frame would be Security and Defense since the action is taking place on the border. All terrorist attacks are coded as Security and Defense, but attention should be paid to potential criminal, legal, or any other aspects and double coded accordingly.

\paragraph{8. Health and safety:}
The potential health and safety outcomes of any policy issue (e.g. health care access and
effectiveness, illness, disease, sanitation, carnage, obesity, mental health infrastructure and
building safety). Also policies taken to ensure safety in case of a tragedy would fit under this
(e.g. emergency preparedness kits, lock down training in schools, disaster awareness classes for
teachers).
It includes any discussion of the various capital punishment methods and procedures and any mentions of refugees. Often used in conjunction with Quality of Life.

\paragraph{9. Quality of life:} The benefits and costs of any policy on quality of life. The effects of a policy on people’s wealth, mobility, access to resources, happiness, social structures, ease of day-to-day routines, quality of community life, etc. It includes any mention of people receiving generic “benefits”, adoptions, and weddings. Often used in conjunction with Health and Safety.

\paragraph{10. Cultural identity:}
The social norms, trends, values and customs constituting any culture(s).
It includes: i) language issues and language learning; ii) patriotism and national traditions, the
history of an issue or the significance of an issue within a group or subculture; iii) census and
demographics; iv) cultural shifts in a group or society; v) cultural norms of ethnic and political
groups. May also include stereotypes or assumed preferences and reactions of a group (e.g., an
affinity for Republicans to wear cowboy hats); vi) references and quotations of famous people
like politicians, leaders or representatives of a subculture.

\paragraph{11. Public opinion:} The opinion of the general public. It includes references to general social attitudes, protests, polling and demographic information, as well as any public passage of a proposition or law (i.e. “California voters passed Prop 8”). All the opinions that represent the sentiment of a group will be coded as Public opinion. However, a group of experts in a particular domain gets coded according to their domain (e.g. police officers in Crime and Punishment, or climate scientists in Capacity and Resources).

\paragraph{12. Political:}
In general, any political considerations surrounding an issue.
It includes political actions, maneuvering, efforts or stances towards an issue (e.g. partisan
filibusters, lobbyist involvement, deal-making and vote trading), mentions of political entities or
parties (e.g., Democrats, Republicans, Libertarians, Green Party). When a headline mentions
“both sides” this refers to politics.

\paragraph{13. Policy prescription and evaluation:} The analysis of whether hypothetical policies will work or existing policies are effective. What is/isn’t currently allowed and what should/shouldn’t be done? “Policy” encompasses formal government regulation (e.g., federal or state laws) as well as regulation by businesses (e.g., sports arenas not allowing the sale of alcohol). This frame dimension—perhaps more than any other—is likely to appear frequently across texts. Yet care should be given to only use this code category as the primary frame when the main thrust of an article is really about policy, for example when it describes the success and failure of existing policies or proposes policy solutions to a problem.

\paragraph{14. External regulation and reputation:} In general, the country’s external relations with another nation; the external relations of a state with another.This frame includes: i) trade agreements and outcomes; ii) comparisons of policy outcomes between different groups or regions; iii) the perception of one nation, state, and/or group byanother (for example, international criticisms of the United States maintaining capital punishment); iv) border relations, interstate or international efforts to achieve policy goals; v) alliances or disputes between groups.

\paragraph{15. Other:} Any frame signal that does not fit in the first 14 dimensions.

\subsection{Further results}
\label{app:results}

\begin{figure}[h!]
    \centering
    \includegraphics[width=1\linewidth]{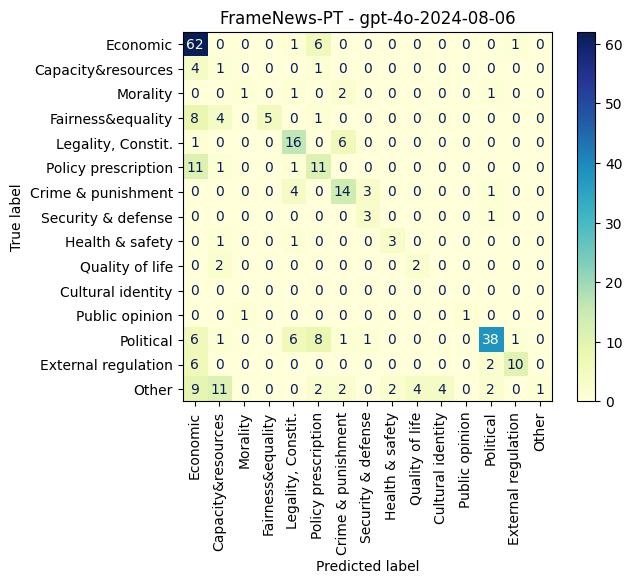}
    \caption{Confusion matrix of the best prompt and best model (chatGPT-4o) in comparison with annotator 1.}
    \label{fig:enter-label}
\end{figure}

\begin{table*}[h!]
\begin{minipage}{1\linewidth}
    \begin{subtable}{\linewidth}
    \centering
    \small
        \begin{tabular}{l>{\columncolor{gray!20}}lll>{\columncolor{gray!20}}l}
        \toprule
        \textbf{Model} & \textbf{Global F$_1$} & \textbf{F$_1$ (ann 1)} & \textbf{F$_1$ (ann 2)} & \textbf{Accuracy} \\
        \midrule
        gpt-4o-2024-08-06        & \textbf{0.50} $_{\pm 0.04}$ & 0.51 $_{\pm 0.04}$ & 0.49 $_{\pm 0.03}$ & \textbf{0.59} $_{\pm 0.02}$ \\
        Qwen2.5-7B-Instruct      & 0.45 $_{\pm 0.04}$ & 0.46 $_{\pm 0.04}$ & 0.43 $_{\pm 0.03}$ & 0.54 $_{\pm 0.02}$ \\
        Llama-3.1-8B-Instruct    & 0.36 $_{\pm 0.04}$ & 0.37 $_{\pm 0.03}$ & 0.34 $_{\pm 0.04}$ & 0.44 $_{\pm 0.04}$ \\
        gemma-3-4b-it            & 0.35 $_{\pm 0.03}$ & 0.36 $_{\pm 0.02}$ & 0.34 $_{\pm 0.04}$ & 0.43 $_{\pm 0.02}$ \\
        Qwen2.5-1.5B-Instruct    & 0.29 $_{\pm 0.09}$ & 0.30 $_{\pm 0.09}$ & 0.28 $_{\pm 0.08}$ & 0.41 $_{\pm0.08}$ \\
        Qwen2.5-3B-Instruct      & 0.33 $_{\pm 0.08}$ & 0.34 $_{\pm 0.08}$ & 0.32 $_{\pm 0.07}$ & 0.40 $_{\pm0.07}$ \\
        Llama-3.2-3B-Instruct    & 0.11 $_{\pm 0.07}$ & 0.12 $_{\pm 0.06}$ & 0.10 $_{\pm 0.07}$ & 0.26 $_{\pm0.06}$ \\
        Llama-3.2-1B-Instruct    & 0.16 $_{\pm 0.02}$ & 0.17 $_{\pm 0.02}$ & 0.15 $_{\pm 0.02}$ & 0.24 $_{\pm0.03}$ \\
        \bottomrule
        \end{tabular}
    \caption{\texttt{FrameNews-PT} zero-shot results. We report F$_1$ scores averaged on both annotators (global F$_1$), against annotator 1 (ann1) and annotator 2 (ann 2).}
    \end{subtable}
    
    \vspace{1em}

    \begin{subtable}{1\linewidth}
    \centering
    \small
        \begin{tabular}{l>{\columncolor{gray!20}}l>{\columncolor{gray!20}}l}
        \toprule
        \textbf{Model} & \textbf{Global F$_1$} & \textbf{Accuracy} \\
        \midrule
        gpt-4o-2024-08-06     & \textbf{0.47} $_{\pm 0.01}$ & \textbf{0.46} $_{\pm 0.01}$ \\
        Qwen2.5-7B-Instruct   & 0.39 $_{\pm 0.02}$ & 0.40 $_{\pm 0.01}$ \\
        Llama-3.1-8B-Instruct & 0.35 $_{\pm 0.02}$ & 0.37 $_{\pm 0.02}$ \\
        gemma-3-4b-it         & 0.34 $_{\pm 0.01}$ & 0.37 $_{\pm 0.03}$ \\
        Qwen2.5-3B-Instruct   & 0.30 $_{\pm 0.05}$ & 0.30 $_{\pm 0.04}$ \\
        Llama-3.2-3B-Instruct & 0.16 $_{\pm 0.12}$ & 0.27 $_{\pm 0.09}$ \\
        Qwen2.5-1.5B-Instruct & 0.21 $_{\pm 0.12}$ & 0.23 $_{\pm 0.11}$ \\
        Llama-3.2-1B-Instruct & 0.06 $_{\pm 0.06}$ & 0.12 $_{\pm 0.07}$ \\
        \bottomrule
        \end{tabular}
        \caption{\texttt{MFC} zero-shot results. We report scores against the gold label.}
    \end{subtable}
    \caption{Zero-shot results with generative models on both datasets. Accuracy is calculated by considering the output as correct if it matches at least one annotation. The standard deviation is calculated based on the three prompt templates used to evaluate each model. Best results are \textbf{in bold}.}
    \label{tab:all_results}
    \end{minipage}
\end{table*}

\end{document}